\documentclass[conference]{IEEEtran}
\usepackage{cite}
\usepackage{amsmath,amssymb,amsfonts}
\usepackage{algorithmic}
\usepackage{graphicx}
\usepackage{textcomp}
\usepackage{xcolor}
\usepackage{tabto}
\def\BibTeX{{\rm B\kern-.05em{\sc i\kern-.025em b}\kern-.08em
    T\kern-.1667em\lower.7ex\hbox{E}\kern-.125emX}}
\begin{document}

\title{Print Error Detection using Convolutional Neural Networks}

\author{

\IEEEauthorblockN{Suyash Shandilya}
\IEEEauthorblockA{\textit{Research Fellow} \\
Defence Research and Development Organisation\\
New Delhi, India\\
su.sh2396@gmail.com}

}

\maketitle

\begin{abstract}
This paper discusses the need of an automated system for detecting print errors and the efficacy of Convolutional Neural Networks in such an application. We recognise the need of a dataset containing print error samples and propose a way to generate one artificially. We discuss the algorithms to generate such data along with the limitaions and advantages of such an apporach. Our final trained network gives a remarkable accuracy of 99.83\% in testing. We further evaluate how such efficiency was achieved and what modifications can be tested to further the results.
\end{abstract}

\begin{IEEEkeywords}
Convolutional neural networks, Print errors, Optical Character Recognition, Reprography
\end{IEEEkeywords}

\section{Introduction}
Reprography has been among the oldest industries in modern times. Even in the age of digitisation, printed documents provide a special advantage in security and ease. In institutions where security of data is foremost, an entire unit of machines and experts is devoted to perpetual management and distribution of such information.\\
Security concerns often entail criticality of data as well. Our problem originates in the distribution of printed data. Despite having the best systems installed, one cannot rely entirely on the machines to print immaculately. When such errors disrupt even a single character of a large key, the entire key is rendered unusable. Thus a good amount of time and resources are devoted to examining every sheet. Our objective here is to save the capital and time invested in this process by automating this explicit evaluation. With the advancement in technologies like Convolutional Neural Networks, we successsfully attempted at designing a ConvNet which proved to be exceptionally efficient in recognising print errors on a large set of data.\\
In a place where more than hundreds of thousands of sheets of printed text is produced and processed, collecting sufficient erroneous and commensurate proper text should be a cakewalk. But the foremost priority, of any data generation centre, is privacy. There are strict (and obvious) strictures to disallow access to original data, let alone digitisation. Yet the manually scrutiny every page of the final print is an unavoidable process as even a single error in a sheet might impare the entire data. This is the most time consuming and human intensive part in the entire pipeline.\\
Thus the bottleneck of the problem was to train a network, without any data. This lead us to isolatedly recreate the print process, study and simulate the errors and then train the network.

\paragraph{ \textbf{F} vs. \textbf{E} error}
Perhaps the gravest reprographic pathology that can jeopardise cipher key management, is a print error that makes a character look like another character. Say an LSE (see section) near the bottom of a text makes an 'E' look like an 'F', or a 'J' look like an 'I'. Any system built around the concept of recognising characters, be it Optical Character Recongnition (OCR) \cite{b3},  K nearest neighbors \cite{b5}, Support Vector Machines \cite{b4} - will most likely misread the character, perhaps with a high character confidence. We address this set of problems as F vs E problem. Even manual scrutiny is very prone to such aberrations. To identify such errors correctly, finer character analysis and more intelligent classification is required. Since the objective of our problem is to simply detect an error, a model designed to detect error will work way better than using character recognition and differentiating errors on the basis of confidence.

\subsection{Types of errors}

\begin{figure*}[t!]
\centering
  \includegraphics[scale=1.4]{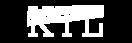}
  \includegraphics[scale=1.35]{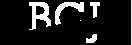}
  \includegraphics[scale=1]{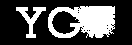}
  \caption{The three types of errors (L to R): Line Print Error(LPE); Line Skip Error(LSE); and Blot Error(BE).}
  \label{fig:differrors}
\end{figure*}

3 distinct types of errors were chosen to be simulated: line print error (LPE), line skip error (LSE) and blot error (BE). Fig \ref{fig:differrors} shows an example of the three types. These might very conspicuous errors but they are actually very small if you consider the ratio of the sample segment to the entire paper area (\textless 0.1\% in pixel count). Having said that, only those errors are being considered which make the text seriously ambiguous. Most of the original errors were wither more benign or more charecterisable than the simulations. This has been chosen so to make the detection model more robust to changes and less sensitive to innocuous aberrations.(See Fig \ref{fig:differrors})

\paragraph{Line Print Error}\label{LPE}

Erratic print head movements can often create line marks. Impurities like dust on the paper often adsorb and drag fresh ink creating similar artefacts. The simulated errors have been made fuzzier than what was observed to entail more irregularities. Almost all LPEs we observed were solid lines.

\paragraph{Line Skip Error}\label{LSE}

Mathematically, it is simply the binary opposite of the LPEs. At times the printhead misses parts of the paper, omitting relevant regions creating errors as shown in the 2\textsuperscript{nd} image in Fig \ref{fig:differrors}. Similar to LPEs, they are slightly exaggerated and fuzzier. As we will see later, the rare instances of misclassification are mostly dominated by these fuzzy LSEs. (Fig \ref{fig:finalerrors})

\paragraph{Blot Error}\label{BE}

A very common and pertinent error across all printing press is the inkblot error. Though having simple, explainable physical origins, they are the most random in design among all the observations. Capturing their randomness as closely as possible required significant heuristics and computational capacity.

\subsection{ConvNets and other prospective solutions}

\begin{figure}[h!]
\centering
  \includegraphics[scale=0.7]{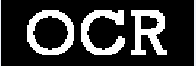}
  \includegraphics[scale=0.77]{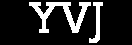}
  \caption{"OCR" (left) can be unambiguously read by an OCR as the characters are well separated. However, "YVJ" (right) often characterises 'YV' as a single character as they are conjoined on the top.}
  \label{fig:goodtext}
\end{figure}

Despite being a pervasive problem across any scale of print production, the industries have till date mostly relied on manual labour as described in the introduction. Therefore not much published work regarding this particular problem statement came across us during our research.\\
OCR usually seems as an obvious solution. Apart from the previously stated issues, its results were very poor because of the overlaping of characters. It may be because of font, or an erroneous print. As can be observed in Fig. \ref{fig:goodtext}, the characters are well readable. Yet many OCRs still bing the 'YV' as a separate character, which in our case would be classified as an error. More importantly, standard OCRs like Tesseract \cite{tesseract} and MATLAB\textsuperscript{\textcopyright}  OCR takes much more time in drawing an inference than a ConvNet.\\

Most of the image processing solutions we tried were significantly slow. We even considered methods like increasing character spacing but none seemed implementable other than ultimately using heuristic/AI solutions. Using frequency transforms to observe prominent spacing was also considered but the scans are huge (3500 X 4200, 45MB) and transforming will take time and not a very exact solution.\\
Neural Networks have recently been acknowledged for their fast \cite{b6} and efficient \cite{b7} abilities in feature detection. Given the scale of our operation, both of these traits are an undeniable virtue. Their large data requirements are easily met as we are generating data ourselves.
Readers may note that we segment our images first to extract the text regions. The reasoning behind this explained in detail in section \ref{preproc}.\\
Our major contributions are: 1) Creating algorithms and a public database \cite{b1} for printing errors which seems to be a first. 2) Establishing the exceptional capacity and efficient implementation of ConvNets in detecting such errors.\\
Future work may deal in testing other modalities like SVM, decision trees and clustering.

\section{Data Generation}

As previously stated, due to the obvious hardships, we could not scan or store them anywhere digitally. So we made a huge synthetic dataset using random sets of alphabets as text. This became the ‘good’ dataset. We then designed algorithms to simulate the observed errors and added the errors to the good dataset. This was labelled as the ‘bad’ dataset. We discuss the designing of the errors here and the basic preprocessing in the next section.

\subsection{Algorithm for LSEs and LPEs}\label{dg}
The algorithm for the 2 types are same except for the 'ink constant' which can only be 0 (LSE) or 255 (LPE). Here, the wedge will refer to the complete artefact which is essentially a bundle of lines drawn close, randomly and with very little variance in slope.\\
The algorithm chooses a primary seed point on the image (it is restricted to pick a point only from the inner 80\% region, uniformly at random). With a variance proportional to the expected girth of the wedge, two normally distributed vectors containing the coordinates of the secondary seeds were built and stored. These secondary seeds become the point of origin of the singular lines that bundle together to form the wedge. The number of lines to be drawn should also be kept proportional to the expected wedge girth. Another Gaussian random vector for the slope of the line is used to store the angle of slope for each line. The mean is chosen randomly between 0 and $\pi$. Subsequently, a simple loop is iterated across all the secondary seed coordinate pairs, and their respective slopes.\\
The reader may note that the random vector hear is the angle of slope and not its tangent. This means the same variance in angle will produce relatively parallel horizontal lines, but the vertical wedge will be more diffused in shape. This led us to explicitly add images with even lower variance and mean angle around $\pi$/2. This also improved the accuracy further, the reasons for the same shall be discussed in the section \ref{randa}.

\subsection{Algorithm for BE}
The key idea is to add random lengths on the edges of a solid circular manifold. Initially, uniform random values seemed to produce credible spots. But when the size (variance) of the blot was increased, pertinent curves started to appear because of rounding errors and low number of pixels. Fearing the ConvNet might train to read those curves specifically, the distribution was changed to normal. It produced slightly splashier spots than uniform but it also entailed more randomness.
Only a single primary seed point needs to be chosen uniformly at random on the image. A simple loop was iterated from 0 to 2$\pi$ to draw rays whose lengths are normally distributed about a mean (radius of the spot) and a variance (degree of ‘splash’). The density of the spot can be controlled by choosing the discrete differential angle between each iteration. In all the algorithms the non-linearity due to rounding error adds to the verisimilitude of the simulation.\\

A general caveat is in order before deploying these algorithm for creating a large dataset. Randomness generated by computer simulations have their limitations. They are more pronounced when the number of pixels is low and there is discretization involved. More the number of datapoints, more a feature would be repeated. This might lead to a loss of generalization, resulting in poor test results. There is a good scope in the algorithm to further randomize it and it is advisable to exploit it before reproducing the result on a larger dataset.

\section{Data Preprocessing}\label{preproc}

The scanners we deployed scanned the images as RGB files in TIFF even when operated in grayscale mode. Since the nature of our target features is binary, the data was naturally converted to grayscale and binarized. Binarization removes redundant errors from scanning in preprocessing itself \cite{b8}. Since different character combinations make different length words, the segments were finally scaled to the standard size of 132 X 45 pixels.\\
We have still not stated the reason behind choosing to train and process the sheets as segmented texts rather than as whole sheets. Although mostly intuitive, the immense advantages that segmentation provides is worthy of addressing.\\
First, we do not have access to full original sheets. Reproducing them strictly will require more time than simply reproducing the text reliably. This means more data can be generated in less time. Second, error simulation is more efficient when being randomly applied on pure text segments than full sheets. Third, our model will not generate false alarm when an unwanted artefact appears on a blank area. This hones our concern on the unreadability of our text. Fourth, since the datapoints are now small in size, which yields a smaller network model. It is only trained to detect print errors on text and will not be overtrained by irrelevant data. Thus it will focus on the print errors more intricately, delivering better results in critical problems like F vs E. Fifth, smaller networks are much faster and lighter than larger ones \cite{b9}. This would multiply the rate of processing many fold if we deploy them parallelly\cite{b11}.\\
\begin{figure*}[!t]
\centering
  \includegraphics[scale=0.7]{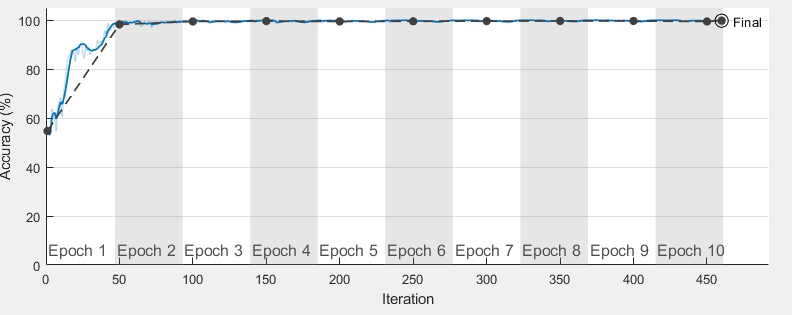}
  \caption{Training process during the final model. The black dashed line shows the validation performance. In the final validation test, the result was \textbf{99.85\%} - slightly more than the claimed accuracy. }
  \label{fig:trainingprogress}
\end{figure*}
\section{CNN Implementation}

To the best of our knowledge, this is the first application of CNNs in print error detection. Since no pretrained network existed, we created our own architecture and trained it accordingly. Interestingly, this problem appeared to be of the nature where a shallow network completely outperforms a deep one\cite{b10}. This is primarily because of the fact that the data points are too small to propagate sensibly beyond 3 convolution+activation layers. Intuitively as well, one can observe that the features are very discernible as is, and there’s little high level inference required. The final implementation is described below while the reasoning and approach will be briefly discussed in the next section.\\
The input image size – as previously mentioned – is 45 X 132 X 1 (single channel, i.e. grayscale image). Only 2 layers of 2D convolution having 3 and 5 output channels respectively separated by a full stride 3X3 pooling layer were used. Batch normalisation was used after each convolution layer. The final image was then flattened and funnelled to a 2 node dense layer via a 100 node dense layer. All the relevant layers used ReLU activation function \cite{b14} except for the final 2 node dense layer which used softmax activation function \cite{b15}. Crossentropy function was used to calculate the cost of misclassification in the final output layer\cite{b16}. No dropout\cite{b20} was used in the final architecture.\\
From the dataset of 20000 images, 12000 were used to train the model, 6000 were used in testing, while the remaining were used to validate the training after every 50 iterations. Stochastic Gradient descent with momentum \cite{b18} was used as the solver. The learning rate and momentum were kept constant at 0.1 throughout the training. Nominal L2 regularization (10\textsuperscript{-4}) was used to limit overfitting\cite{b19}. Each iteration was performed on a minibatch of 256 images. The full training proceeded for 10 epochs.\\

\section{Results and Analysis}\label{randa}
\begin{figure}[h]
\centering
  \includegraphics[scale=0.35]{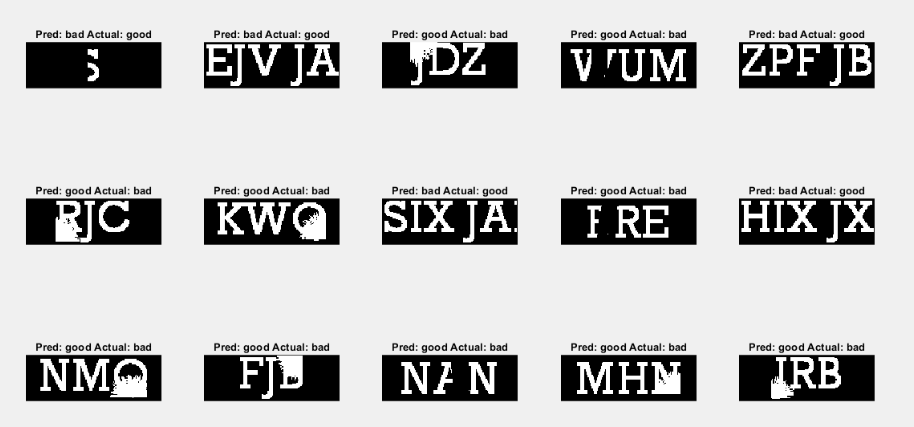}
  \caption{Misclassification cases when square filters were used in a typical case. Observe the prominence of BEs and their location on the text segment. }
  \label{fig:olderrors}
\end{figure}

A staggering accuracy of \textbf{99.83\%} in testing was achieved with the final architecture. In this section, we will elaborate our experimentation with different architectures and parameters to deduce how the network drew its inference. In the next few paragraphs we will be discussing what training options and architectures were tried before finalising the one described in the previous section.\\
Stochastic Gradient Descent with Momentum (SGDM) was the foremost preference for the solver. It is known to perform better than its relatively newer competitors like ADAM or RMSprop but usually has slower convergence \cite{b2}. The training time for our network was less than 10 minutes in the worst case and about 5 minutes in an average case for 10 epochs. Since performance was much more imperative than training time in our case, we pertained with SGDM\cite{b17}. As can be observed from the training progress in Fig \ref{fig:trainingprogress}, the network learns the discriminating manifold within the first few epochs itself. The validation results corroborate that the learning is in fact generalised and not regurgitated. Not by a big margin, the accuracy was nevertheless reduced in either increasing or decreasing the number of epochs. About 0.2\% increase was pertinent on increasing the minibatch size from 128 to 256 across all other parametric variations.\cite{mbsize} Other parameters showed little significant correlation with testing accuracy.\\
As mentioned earlier, the shallower architectures proved to be more efficient than their deeper counterparts. There was a significant difference in using 2 and 3 convolutional + activation layers, former being the better performer. Using ReLU activation with 2D convolution is a standard preference and stood its reputation in our implementation as well. LeakyReLU \cite{b13} did not provide any advantage in accuracy.\\
Initially both the convolutional layers had the same square 5X5 filters. It gave a splendid accuracy in the range of 98.88 to 99.15\%. 
\begin{figure}[h]
\centering
  \includegraphics[scale=0.35]{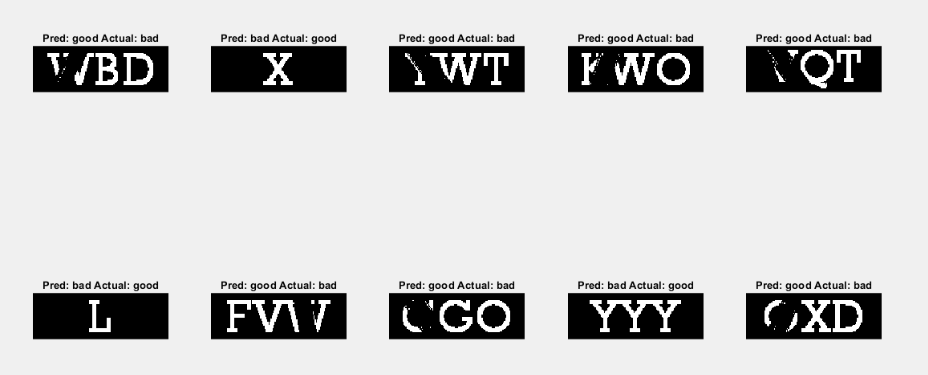}
  \caption{Misclassifications after using rectangular filters and additional LSE errors. There are no BEs. Observe how all the misclassified errors are unambiguously readable (except for one) and are fuzzy in nature. Compare this with Fig. \ref{fig:olderrors}}
  \label{fig:finalerrors}
\end{figure}
It was observed (as shown in Fig \ref{fig:olderrors}) that the rare cases of misclassification only included those BE samples where the blot occurred on a corner. (It may be noted that the training data consists of about 5000 BE samples and the misclassification occurred in less than 10 such cases in the worst of trials). An intuitive modification was to use non-square filters as the text is mostly dominated by prominently vertical and horizontal features while the ‘blots’ would not show such characterization. Proceeding on these lines, the 2 layers were then made to have 5X10 and 10X5 sized filters. This shot up the accuracy to more than 99.5\%. It can be seen that the BE misclassification cases were eliminated virtually entirely (compare fig and fig).\\
Another observation that was drawn from Fig \ref{fig:olderrors} was about the LSEs. It was noted (in multiple tests) that the misclassified LSEs were almost entirely dominated by artefacts that were solid (less fuzzy) and very vertical. Intuitively as well, it is an efficient way to blemish printed text. This was in part because there were relatively few solid, vertical LSEs (see section \ref{dg}). And because they blend excellently with the text making them harder to categorise. This caused us to tweak our algorithms to create more such artefacts for training. We made 1000 such errors and added them to our dataset. The result was as expected and all cases of such misclassification disappeared. Interesting thing to note from this expedition was that our methods (algorithms, per se) were malleable enough to specially accomodate any such special, pertinent, artefact which may sabotage the readability of the document.\\
The most unsurprising yet fascinating improvement was owed to Batch normalization \cite{b12}. Without it, both the training and testing performance were very incompetent and erratic. The best case accuracy was 89\% which is practically unacceptable in our application. Even during training, the accuracy would fluctuate heavily across epochs.\cite{b19}\\
In our prior analysis, we briefly considered experimenting with asymmetric cost function in later stages of implementation \cite{asymcostfn}. This was synchronous with the practical expenses in implementation – the cost of printing few sheets of paper redundantly is much less than printing a single sheet which may be ambiguous to read. However, we did not implement it at any stage as the accuracy was very high. Even a slight asymmetry in classification would wreak havoc on the testing result. Since these networks analyse multiple text segments on a single sheet – which are typically in the range of 100 to 300 – the error per sheet would seriously increase in any such implementation. We do however hope to try it on other databases that might be available as more research in this field is published in public domain.\\

\section{Conclusion and future work}

Our aim was to automate the print error detection process in a large scale printing press where data readability is of utmost importance. We evaluated multiple solutions and realised that ConvNets would be the most efficient and fast way to tackle this problem. We discussed the most pertinent categories of print errors and how can they be simulated. Our first major contribution was to create a public database \cite{b1} for such a problem which thus far, seemed unavailable. We then established how ConvNets proved to be the optimum solution because of thei speed, and a staggering accuracy of \textbf{99.83\%}. We analysed the heuristics in the implementation and training of network. Another intersting observation is that none of the erroneous images which were misclassified are ambiguos in reading. It shows that despite having a shallow architecure, the model learnt in part, a sense of 'readability' rather than shear feature detection\cite{shallownet}.\\
Our future work would be centred primarily around testing the results on real data and on different kinds of text data. Different text formats will have different features, thus our network may not be a panacea for the entire class of print error detection. Thus far we only worked on uncompressed images and only tested ConvNets.\\

\end{document}